\documentclass[sigconf,review]{acmart}
\makeatletter
\let\roughcutOriginalLineCount\ACM@mk@linecount
\renewcommand{\ACM@mk@linecount}{%
  \roughcutOriginalLineCount
  \sbox{\ACM@linecount@bx}{\phantom{\usebox{\ACM@linecount@bx}}}%
}
\makeatother
\usepackage{enumitem}
\usepackage{multirow}
\usepackage{tabularx}
\usepackage{colortbl}
\usepackage{listings}
\usepackage{tcolorbox}
\tcbuselibrary{breakable,listings,skins}

\lstdefinestyle{agentprompt}{
  basicstyle=\ttfamily\fontsize{6.6pt}{7.3pt}\selectfont,
  breaklines=true,
  breakatwhitespace=false,
  columns=fullflexible,
  keepspaces=true,
  showstringspaces=false,
  tabsize=2,
  aboveskip=0pt,
  belowskip=0pt
}

\tcbset{
  agentprompt/.style={
    enhanced,
    breakable,
    listing only,
    listing engine=listings,
    listing options={style=agentprompt},
    colback=black!1,
    colframe=black!45,
    colbacktitle=black!8,
    coltitle=black,
    boxrule=0.45pt,
    arc=1pt,
    left=1.5mm,
    right=1.5mm,
    top=1mm,
    bottom=1mm,
    fonttitle=\bfseries\small,
    before skip=6pt,
    after skip=8pt
  }
}
\AtBeginDocument{%
  }

\setcopyright{acmlicensed}
\copyrightyear{2027}
\acmYear{2027}
\acmDOI{}
\acmISBN{}
\acmConference[KDD '27]{ACM KDD 2027}{2027}{San Jose, California}

\begin{document}


\title{RoughCut-Bench: A Benchmark for Automatic Video Rough-Cut with Large Vision-Language Models}

\author{Ruisi Zhao}
\affiliation{%
  \institution{ReLER, CCAI}
  \institution{Zhejiang University, ByteDance}
  \city{Hangzhou, Zhejiang}
  \country{China}
  }
\email{zhaors00@zju.edu.cn}

\author{You Li}
\affiliation{%
  \institution{ReLER, CCAI}
  \institution{Zhejiang University, ByteDance}
  \city{Hangzhou, Zhejiang}
  \country{China}
  }
\email{uli2000@zju.edu.cn}

\author{Sihan Chen}
\authornote{Project leader.}
\affiliation{%
  \institution{ByteDance}
  \city{Beijing}
  \country{China}
  }
\email{sihan.chensihan@gmail.com}

\author{Aosong Cheng}
\affiliation{%
  \institution{Peking University, ByteDance}
  \city{Beijing}
  \country{China}
  }
\email{2301210197@stu.pku.edu.cn}

\author{Chixiang Ma}
\affiliation{%
  \institution{ByteDance}
  \city{Beijing}
  \country{China}
  }
\email{chixiangma@gmail.com}

\author{Sen Pei}
\affiliation{%
  \institution{ByteDance}
  \city{Beijing}
  \country{China}
  }
\email{senpei.ai@gmail.com}

\author{Zhen Yang}
\affiliation{%
  \institution{ByteDance}
  \city{Beijing}
  \country{China}
  }
\email{yz\_tencent\_person@163.com}

\author{Hehe Fan}
\affiliation{%
  \institution{School of Artificial Intelligence}
  \institution{Zhejiang University}
  \city{Hangzhou, Zhejiang}
  \country{China}
  }
\email{hehefan@zju.edu.cn}

\author{Dongliang He}
\affiliation{%
  \institution{ByteDance}
  \city{Beijing}
  \country{China}
  }
\email{hedlcc@126.com}

\author{Yi Yang}
\authornote{Corresponding author.}
\affiliation{%
  \institution{ReLER, CCAI}
  \institution{Zhejiang University}
  \city{Hangzhou, Zhejiang}
  \country{China}
  }
\email{yangyics@zju.edu.cn}

\renewcommand{\shortauthors}{Zhao et al.}

\begin{abstract}
Video rough-cut is a core stage in media production workflows, organizing raw media materials into an initial video cut. Despite recent advances in large vision-language models (LVLMs) for video-related tasks, automatic rough-cut remains underexplored, without a comprehensive benchmark for evaluation or mature workflows for producing executable cut plans. To address the evaluation gap, we introduce \textbf{\textit{RoughCut-Bench}}, which covers two practical settings: (1) a \textbf{visually driven setting} over video cases, covering highlight detection and selection, source-video ordering, and adherence to user requirements when provided; and (2) a \textbf{speech-driven setting} over automatic speech recognition (ASR) transcripts, covering filler removal, expression deduplication, and low-value content trimming. Across these settings, the benchmark contains 101 visually driven cases, including 831 raw videos with 1,476 highlight annotations. It also includes 84 speech-driven cases, each consisting of an ASR transcript with annotated cleanup spans. The evaluation compares cut plans with these annotations in both settings and additionally uses a human-aligned agentic evaluator to assess rendered-cut quality in the visual setting. To apply LVLMs to automatic rough-cut, we further propose agentic workflows that guide them to produce structured cut plans from input materials. Experiments show that current LVLMs remain limited in producing expert-aligned cut plans. Our benchmark, code, and prompts are released at \url{https://whalesong-zrs.github.io/RoughCut-Bench-Project-Page/}.

\end{abstract}




\maketitle

\section{Introduction}
In recent years, creating video content like short-form personal vlogs or talking-head videos has become increasingly popular on social media platforms. Processing multiple raw materials into such videos typically begins with \textbf{\textit{video rough-cut}}, which involves selecting segments and arranging them into a coherent cut. The initial cut provides the basis for later fine-cut refinements such as adding captions, background music, and transitions. Despite its foundational role, the manual rough-cut process requires creators to review extensive raw content and carefully construct cut plans, motivating the development of automatic rough-cut.

\par However, systematically studying automatic rough-cut faces two key challenges. \textbf{1) \textit{Lack of comprehensive evaluation.}} Existing benchmarks for large vision-language models (LVLMs) \cite{li2024mvbench, fu2025video, wu2024longvideobench} provide relevant but partial evaluations, focusing on tasks such as temporal grounding and highlight detection \cite{liu2024tempcompass, zala2023hierarchical, zhang2026timelens}, without covering necessary dimensions such as source-video ordering and spoken-content refinement. Moreover, constructing a rough-cut benchmark requires each case to represent a coherent creation scenario rather than a random collection of independent materials, making large-scale collection from web-sourced data difficult. \textbf{2) \textit{Lack of mature executable workflows.}} Recent LVLMs \cite{team2023gemini, achiam2023gpt, yang2025qwen3, guo2025seed1,hong2026glm} have demonstrated strong performance in video understanding \cite{zhang2023video, maaz2024video, li2025videochat} and captioning \cite{xu2024pllava, yang2023vid2seq, islam2024video}, but
existing approaches \cite{chen2024videollm, wang2024lave} generally focus on individual component capabilities, without integrating their outputs into executable cut
plans. Automatic rough-cut instead requires compositional decisions across highlight detection and selection, source-video ordering, and spoken-content refinement. It may also involve multiple source videos and heterogeneous input types, requiring both long-context reasoning and support for different modalities.
\par To this end, we introduce \textbf{\textit{RoughCut-Bench}}. It covers two practical settings: (1) a visually driven setting over video cases, covering highlight detection and selection, source-video ordering, and adherence to user requirements when provided; and (2) a speech-driven setting over automatic speech recognition (ASR) transcripts \cite{radford2023robust, lou2020end}, covering filler removal, expression deduplication, and low-value content trimming. RoughCut-Bench provides professionally annotated data and setting-specific evaluation protocols. To apply current LVLMs to automatic rough-cut, we further propose agentic workflows that guide them to produce executable cut plans.

\par More specifically, RoughCut-Bench contains 101 visually driven cases comprising 831 raw videos and 1,476 professional highlight annotations. These cases span six broad real-world scenarios, such as outdoor vlogs and daily-life scenes. Each case is annotated with highlight timestamps and a reference source-video order. Among them, 39 cases additionally include user requirements, such as target duration, preferred source-video order, or specified highlights to retain or remove, enabling requirement-constrained evaluation. For the speech-driven setting, RoughCut-Bench contains 84 cases derived from talking-head videos, covering scenarios such as beauty/fashion and lifestyle vlogs. Each case consists of an ASR transcript with annotated spans for spoken-content refinement.

\par \textbf{For comprehensive evaluation}, RoughCut-Bench combines annotation-based metrics and human-aligned evaluation. These metrics compare cut plans with annotations in both settings. For speech-driven rough-cut, this comparison is well suited because its cleanup decisions directly correspond to span annotations. For visual rough-cut, however, annotation-based metrics cannot fully assess cut quality because they evaluate structured cut plans without viewing the rendered video. Moreover, a single professional reference may not represent all reasonable choices in visual rough-cut. We therefore introduce a human-aligned agentic evaluator that assigns each rendered cut a visual quality score while considering user requirements when provided. By comparing the evaluator’s scores and reasons with those provided by professionals, we iteratively refine it to improve alignment with professional judgment.

\par \textbf{For automatic rough-cut}, we propose agentic workflows \cite{yao2023tree, shen2023hugginggpt, yao2022react} that guide LVLMs to produce structured cut plans from input materials. To handle long multi-video contexts, the visual workflow employs a \textit{Video-Understanding Agent} and a \textit{Cut-Planning Agent}. The former identifies candidate highlights and provides corresponding reasons and visual descriptions, while the latter reasons \cite{wei2022chain} across candidate highlights to generate the final cut plan, including highlight selection and source-video ordering. For the speech-driven setting, a \textit{Transcript-Cleanup Agent} analyzes each complete ASR transcript and outputs structured cleanup records, corresponding reasons, and a cleaned transcript.

\par With these designs, we provide a systematic protocol for studying automatic rough-cut. We evaluate representative LVLMs on RoughCut-Bench using the proposed agentic workflows. We also conduct ablation studies to validate the workflow design. We further conduct a professional user study to assess agreement between the evaluator's scores and human scoring trends across models. In summary, our main contributions are as follows:

\begin{itemize}[leftmargin=*]
    \item We formulate automatic rough-cut as a compositional creation task. It requires models to produce executable cut plans from source videos or ASR transcripts.

    \item We establish RoughCut-Bench with visually driven video cases and speech-driven ASR cases, professional annotations, and complementary evaluation protocols.

    \item We propose agentic workflows to guide representative LVLMs on automatic rough-cut. Experiments show that current LVLMs exhibit uneven performance across rough-cut capabilities.
\end{itemize}

\section{Related Work}
\begin{figure*}[t!]   % [t] 表示图片放置在页面顶部
    \centering
    \includegraphics[width=\textwidth]{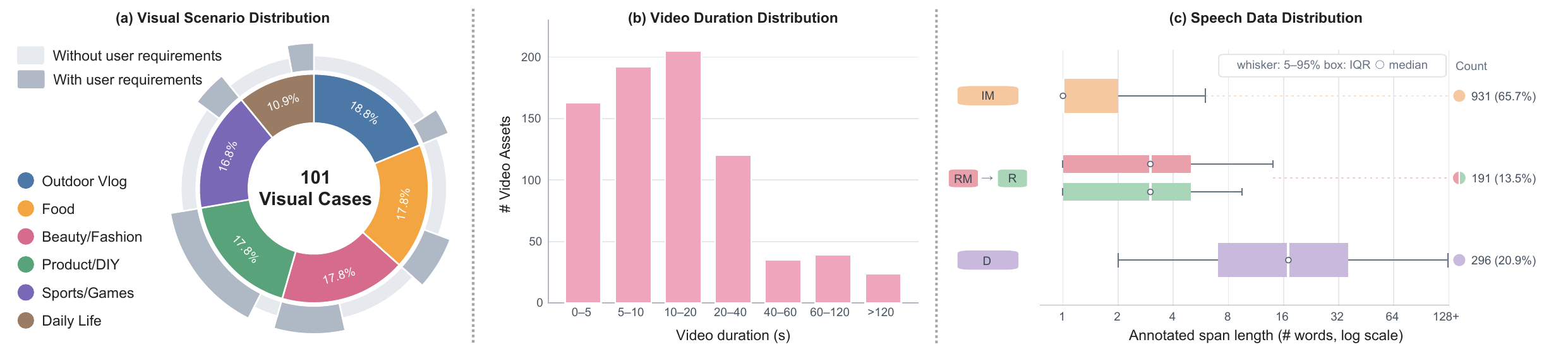} % 图片路径及宽度设置
    \vspace{-0.6 cm}
    \caption{Dataset statistics of RoughCut-Bench (Sec.~\ref{sec:data-construction}). (a) Distribution of the 101 visual cases across six creation scenarios; the outer bars distinguish cases with and without user requirements. (b) Duration distribution of the 831 raw videos. (c) Counts and annotated span-length distributions of the IM, RM$\rightarrow$R, and D speech-cleanup labels.}
    \label{fig:data_analyse} % 标签用于引用
    \vspace{-0.3 cm}
\end{figure*}

\subsection{Video-Language Benchmarks}

Existing benchmarks evaluate models from different perspectives, including general video understanding \cite{li2024mvbench, fu2025video, mangalam2023egoschema, patraucean2023perception}, long-video reasoning \cite{wu2024longvideobench, zhou2025mlvu, wang2025lvbench, rawal2024cinepile}, and temporal perception \cite{liu2024tempcompass}. Most are designed to measure how well models interpret a given video, rather than whether they can organize raw materials into an initial cut.

\par Some benchmarks move closer to video production scenarios by requiring models to localize relevant moments, ground answers in temporal evidence, or understand video-cut operations \cite{xiao2024can, zala2023hierarchical, afouras2023ht, yuan2026momentseeker, gao2026omniground, li2025veu}. These tasks overlap with rough-cut in temporal boundary estimation and highlight selection, but do not jointly evaluate cut planning, adherence to user requirements, spoken-content refinement, and rendered-cut quality. Moreover, comparison with a single reference may not capture all acceptable cut plans. Model-based evaluators provide a scalable approach to assessing such open-ended quality \cite{zheng2023judging, liu2023g, lee2024prometheus, xiong2025llava}, although their reliability requires validation against human judgments.

\subsection{LVLMs for Video-Related Tasks}

Recent LVLMs have achieved strong performance on video-related tasks. They can answer questions about video content \cite{zhang2023video, li2025videochat, maaz2024video, jin2024chat}, produce dense descriptions or summaries \cite{yang2023vid2seq, xu2024pllava, islam2024video}, and reason over long videos with memory mechanisms, token compression, or long-context modeling \cite{song2024moviechat, li2024llama, wang2024internvideo2, shu2025video}. However, rough-cut requires more than recognizing or describing video content: models must select useful moments, estimate temporal boundaries, reason over multiple source videos, and produce executable cut plans.

Temporal grounding, moment retrieval, and highlight detection are closely related to rough-cut because they study timestamp-level localization and important-moment selection \cite{moon2023query, lin2023univtg, ren2024timechat, huang2024vtimellm}. Agentic workflows have also been explored for video understanding \cite{wang2024videoagent, chen2025lvagent, wang2025videotree}, streaming video processing \cite{chen2024videollm, liu2024streamchat}, and multimodal tool use \cite{yao2022react, shen2023hugginggpt, gupta2023visual, suris2023vipergpt}. Nevertheless, they rarely produce executable cut plans across multiple videos. Speech-driven rough-cut is related to ASR-based transcript processing \cite{radford2023robust, lou2020end}, but it requires structured cleanup decisions rather than free-form transcript rewriting.

\section{RoughCut-Bench Suite} \label{sec:benchmark-suite}

\par This section presents the design of RoughCut-Bench. We first introduce the rough-cut task in \textbf{Task Definition
(Sec.~\ref{sec:task-definition})}, then describe the data source, annotation protocol, and dataset statistics in \textbf{Data Construction and Statistics (Sec.~\ref{sec:data-construction})}. Next, we present the \textbf{Evaluation Protocols (Sec.~\ref{sec:evaluation-protocols})}, including the annotation-based metrics and human-aligned agentic evaluator used in the benchmark. Finally, we describe the \textbf{Agentic Workflows for Rough-Cut (Sec.~\ref{sec:execution-workflow})}, which guide different LVLMs to produce structured cut plans from input materials.

\subsection{Task Definition}
\label{sec:task-definition}

\par Automatic video rough-cut requires models to produce an executable cut plan from input materials by removing low-value content, detecting and selecting highlights, and determining an output order under optional user constraints. It is a compositional task that requires coordinated reasoning to construct a cut plan.

\par In RoughCut-Bench, each visually driven case contains one or more source videos. The model detects and selects highlight segments from the inputs and determines the sequence of those videos in the initial cut. The resulting cut plan records the selected highlights in output order, specifying the source video, start time, and end time of each highlight; highlights from the same source video retain their original temporal order. When user requirements are provided, the model must follow them when generating the cut plan. These requirements may specify the target duration, the source-video order, or content that should be retained or removed.

\par For speech-driven rough-cut, we formulate spoken-content refinement as a transcript-based cleanup task using ASR outputs rather than raw audio. This setting captures a practical rough-cut scenario for talking-head materials, where cut plans are primarily determined by spoken content. Models perform filler removal, expression deduplication, and low-value content trimming on each complete transcript. The resulting cleanup decisions can be mapped to source-video intervals using the corresponding ASR timestamps.

\begin{figure*}[t!]   % [t] 表示图片放置在页面顶部
    \centering
    \includegraphics[width=\textwidth]{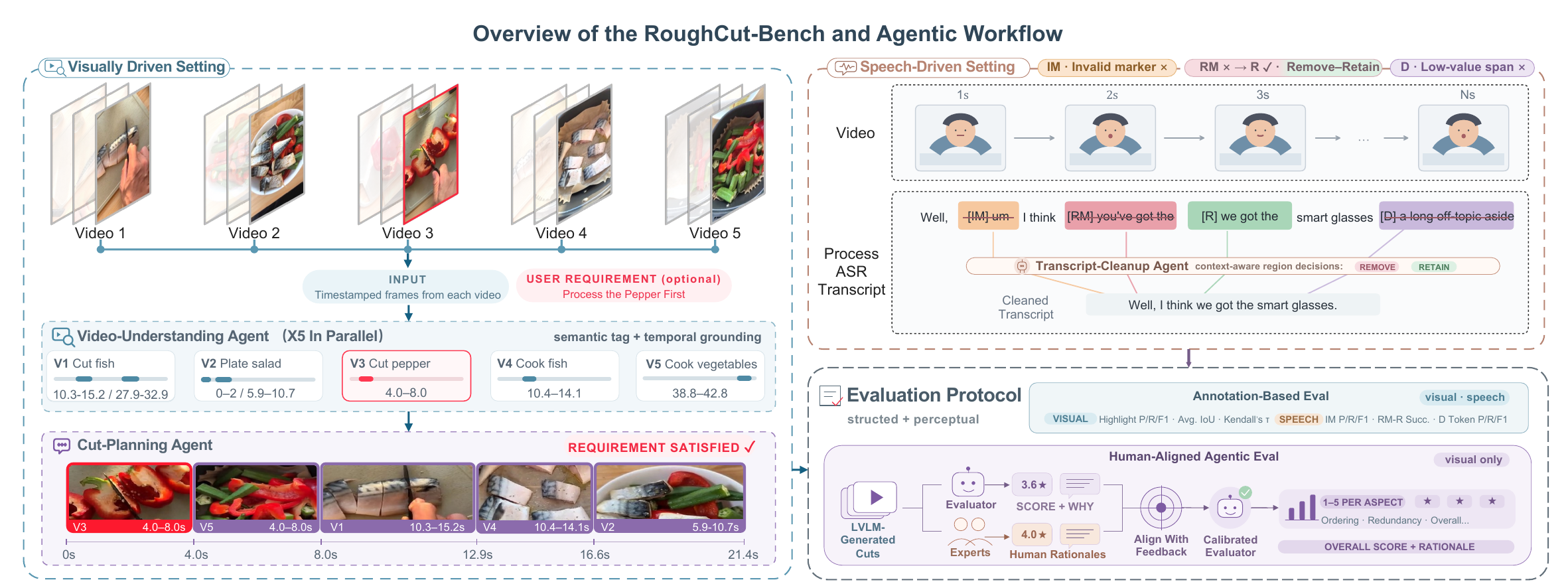} % 图片路径及宽度设置
    \vspace{-0.8 cm}
    \caption{Proposed agentic workflows and complementary evaluation in RoughCut-Bench (Secs.~\ref{sec:evaluation-protocols} and~\ref{sec:execution-workflow}). The visual branch detects candidate highlights and provides corresponding reasons and visual descriptions, then performs highlight selection and source-video ordering. The speech-driven branch processes each ASR transcript and produces structured cleanup records, reasons, and a cleaned transcript. Annotation-based metrics compare cut plans with professional annotations, while the human-aligned agentic evaluator assesses rendered cuts using the input materials and available user requirements.}
    \label{fig:workflow} % 标签用于引用
    \vspace{-0.3 cm}
\end{figure*}

\subsection{Data Construction and Statistics}
\label{sec:data-construction}

\par \textbf{Data Scope and Privacy Review.} RoughCut-Bench is curated from real-world media creation data, covering both visually driven and speech-driven settings. In visually driven cases, the input materials share a common scenario or production intent, providing a coherent context for rough-cut planning. In the speech-driven setting, each case consists of an ASR transcript derived from a talking-head video for spoken-content refinement. All released media, transcripts, annotations, and metadata are cleared for research use. Before release, the data are screened to remove personal identifiers, explicit social handles, and other sensitive information.

\par \textbf{Dataset Statistics.} RoughCut-Bench contains 101 cases in the visual setting, comprising 831 raw videos and 1,476 professional highlight annotations. On average, each visual case contains 8.2 videos with a total duration of 3.15 minutes. Each annotated highlight lasts approximately 3 seconds. These cases span six creation scenarios: outdoor vlogs, food, beauty/fashion, product/DIY, sports/games, and daily life. Among them, 39 cases include user requirements for target duration, content selection, or source-video ordering. A duration requirement may specify a target length or a duration range. A selection requirement identifies which videos or segments should be retained or removed. An ordering requirement specifies which content should appear first or last, an explicit storyline, or a temporal or spatial order. A single prompt may combine multiple requirements, e.g., ``Create a 25-second cut, remove the chili-cutting segment, and place the finished-dish segment first.''
\par The speech-driven setting contains 84 cases derived from talking-head videos, covering fashion tutorials, podcasts and Q\&A videos, and lifestyle vlogs. The ASR transcripts contain three types of cleanup labels: invalid markers (IM), removed-to-retained expression pairs (RM$\rightarrow$R), and low-value deletion spans (D). On average, each transcript contains 438 words, 11.1 IM spans, 2.3 RM$\rightarrow$R pairs, and 3.5 D spans. Figure~\ref{fig:data_analyse} summarizes the visual scenario and video-duration distributions, together with the counts and span-length distributions of the speech-cleanup labels.

\par \textbf{Professional Annotation.} Annotators are multimedia practitioners with experience in content production. Before annotation, they received task-specific guidelines and representative examples. For visually driven cases, annotators label highlight timestamps and a reference source-video order. Highlight annotations focus on informative moments that capture key actions or scene changes, while excluding repetitive footage, low-quality shots (e.g., blurred, shaky, or black-screen content), and low-information shots (e.g., empty scenes without salient subjects). The reference order follows a coherent temporal progression (e.g., daytime before nighttime), spatial continuity (e.g., grouping indoor or outdoor scenes), or narrative progression (e.g., ordering the steps of a cooking tutorial). When user requirements are provided, the annotations follow the specified selection, ordering, and duration constraints.

For the speech-driven setting, annotators use three label categories corresponding to the three cleanup operations. \textbf{\underline{I}}nvalid \textbf{\underline{M}}arkers (IM) correspond to filler removal and mark local filler or discourse markers, such as ``uh'' or ``emm''. \textbf{\underline{R}}e\textbf{\underline{M}}oved expression--\textbf{\underline{R}}etained expression pairs (RM$\rightarrow$R) correspond to expression deduplication: the RM label marks a redundant or erroneous expression to remove, while the R label marks the intended expression to retain, e.g., \texttt{[RM]you've got the,[RM] [R]we got the smart glasses.[R]}. \textbf{\underline{D}}eletion (D) corresponds to low-value content trimming and marks longer spans such as off-topic explanations, repetitive setup, or rhythm-breaking digressions.

In both settings, the cases undergo a two-stage annotation process. Each case is initially annotated by one professional annotator and reviewed in full by a second. The two annotators resolve disagreements over highlight boundaries, source-video ordering, or speech-cleanup labels through discussion. Cases that remain ambiguous after discussion are excluded.

\subsection{Evaluation Protocols}
\label{sec:evaluation-protocols}

\par RoughCut-Bench combines annotation-based metrics with a human-aligned agentic evaluator. The annotation-based metrics compare cut plans with professional annotations, covering highlight detection/selection and source-video ordering in the visual setting, as well as span-level cleanup decisions in the speech-driven setting. These metrics serve as the primary evaluation for speech-driven rough-cut because its cleanup decisions directly correspond to annotated spans. For visually driven rough-cut, however, annotation-based metrics cannot fully assess cut quality because they evaluate structured cut plans without viewing the rendered video. Moreover, a single professional reference cannot represent all reasonable choices in visual rough-cut. We therefore complement them with a human-aligned agentic evaluator that assesses the rendered video while considering available user requirements.

\par \textbf{Annotation-based metrics.} For the visual setting, predicted highlights are compared with annotated highlights within each source video using temporal IoU. We consider a prediction matched when its IoU is at least 0.5 and compute \textit{precision}, \textit{recall}, and \textit{F1} from these matches. \textit{Average IoU} is the mean of each predicted highlight's maximum IoU with any annotated highlight. Source-video ordering is evaluated with \textit{Kendall's $\tau$}, which compares the predicted order with the reference order. For the speech-driven setting, we align each cleaned transcript with the annotated ASR transcript to evaluate removal and retention decisions without requiring an exact string match. For each category $c \in \{\mathrm{IM}, \mathrm{D}\}$, $\mathrm{TP}_c$ counts annotated content correctly removed, $\mathrm{FN}_c$ counts annotated content that the model fails to remove, and $\mathrm{FP}_c$ counts content that is incorrectly removed even though it should remain. We compute:
\begin{equation}
P_c=\frac{\mathrm{TP}_c}{\mathrm{TP}_c+\mathrm{FP}_c},\quad
R_c=\frac{\mathrm{TP}_c}{\mathrm{TP}_c+\mathrm{FN}_c},\quad
\mathrm{F1}_c=\frac{2P_cR_c}{P_c+R_c}.
\end{equation}
For IM, $\mathrm{TP}$ and $\mathrm{FN}$ are counted at the annotated-span level, while $\mathrm{FP}$ is counted at the word level over unannotated content. For D, $\mathrm{TP}$, $\mathrm{FN}$, and $\mathrm{FP}$ are counted at the word level, allowing a partially removed D span to receive partial credit. For RM$\rightarrow$R, a pair is successful only when the expression marked by RM is removed and the expression marked by R is retained. We report the proportion of successful pairs as:
\begin{equation}
\mathrm{RM{\to}R}_{\mathrm{Succ.}} = \frac{N_{\mathrm{succ}}}{N_{\mathrm{pair}}}.
\end{equation}
Here, $N_{\mathrm{succ}}$ and $N_{\mathrm{pair}}$ denote the numbers of successful and annotated RM$\rightarrow$R pairs, respectively.
Visual metrics are averaged across source videos and cases, whereas speech metrics are computed from decision counts aggregated across all cases.

\par \textbf{Human-aligned agentic evaluator.} To assess rendered-cut quality, the evaluator examines the rendered cut using the original videos for context and checks whether any provided user requirements are satisfied. It reports scores for \textit{ordering} and \textit{redundancy}, as well as \textit{requirement following} when applicable, and assigns an \textit{overall visual quality} score with a corresponding reason. To align the evaluator with professional judgment, we adopt an iterative feedback process. LVLM-generated cut plans are rendered into videos, and professional annotators assign each video an
overall visual quality score and provide the reason for that score. In each feedback round, we compare the evaluator's scores and reasons with those provided by professionals and refine its system prompt accordingly. After refinement, the evaluator is fixed for all evaluations.

\subsection{Agentic Workflows for Rough-Cut}
\label{sec:execution-workflow}
\begin{table*}[t!]
\centering
\normalsize
\renewcommand{\arraystretch}{1.05}
\setlength{\tabcolsep}{1.2pt}
\begin{tabularx}{\textwidth}{l|*{5}{>{\centering\arraybackslash}X}|*{5}{>{\centering\arraybackslash}X}}
\hline
\multirow{2}{*}{\textbf{Model\textbackslash Metric}} & \multicolumn{5}{c|}{\textbf{\textit{Visual Set-1}}} & \multicolumn{5}{c}{\textbf{\textit{Visual Set-2}}} \\
\cline{2-6} \cline{7-11}
& {\footnotesize\textbf{Precision}} & {\footnotesize\textbf{Recall}} & {\footnotesize\textbf{F1}} & {\footnotesize\textbf{Avg. IoU}} & {\footnotesize\textbf{Kendall's $\tau$}} & {\footnotesize\textbf{Precision}} & {\footnotesize\textbf{Recall}} & {\footnotesize\textbf{F1}} & {\footnotesize\textbf{Avg. IoU}} & {\footnotesize\textbf{Kendall's $\tau$}} \\
\hline
\rowcolor[rgb]{0.969,0.980,1.000}Gemini 2.5 Pro & 0.467 & 0.556 & 0.508 & 0.437 & \underline{0.556} & \underline{0.344} & 0.341 & \underline{0.342} & \underline{0.354} & 0.654 \\
\rowcolor[rgb]{0.969,0.980,1.000}Gemini 3.0 Pro & \textbf{0.490} & 0.568 & \textbf{0.526} & \underline{0.440} & \textbf{0.565} & \textbf{0.345} & \underline{0.342} & \textbf{0.343} & \textbf{0.358} & 0.692 \\
\hline
\rowcolor[rgb]{0.973,1.000,0.969}GPT-5.4 & 0.443 & 0.597 & 0.509 & 0.428 & 0.512 & 0.265 & 0.301 & 0.282 & 0.315 & 0.610 \\
\rowcolor[rgb]{0.973,1.000,0.969}GPT-5.5 & 0.446 & \textbf{0.604} & 0.513 & 0.420 & 0.542 & 0.311 & \textbf{0.343} & 0.326 & 0.336 & 0.702 \\
\hline
\rowcolor[rgb]{1.000,0.992,0.957}Qwen3.5-Plus & \underline{0.468} & 0.571 & 0.514 & \textbf{0.441} & 0.471 & 0.288 & 0.316 & 0.301 & 0.351 & 0.678 \\
\rowcolor[rgb]{1.000,0.992,0.957}Qwen3.7-Plus & 0.459 & 0.595 & \underline{0.518} & 0.429 & 0.535 & 0.279 & 0.272 & 0.275 & 0.336 & \underline{0.706} \\
\hline
\rowcolor[rgb]{0.984,0.973,1.000}GLM-5V-Turbo & 0.418 & 0.564 & 0.480 & 0.396 & 0.509 & 0.241 & 0.271 & 0.255 & 0.308 & 0.594 \\
\rowcolor[rgb]{0.984,0.973,1.000}Seed-2.0-Lite & 0.451 & \underline{0.598} & 0.514 & 0.424 & 0.539 & 0.292 & 0.339 & 0.314 & 0.326 & \textbf{0.715} \\
\hline
\end{tabularx}
% \vspace{0.2em}
\caption{Annotation-based results for the visual setting (Sec.~\ref{sec:visual-reference-results}). \textit{Visual Set-1} evaluates general rough-cut, while \textit{Visual Set-2} evaluates requirement-constrained rough-cut. The best and second-best scores are shown in bold and underlined, respectively.}
\label{tab:visual-reference-results}
\vspace{-0.8 cm}
\end{table*}

\par \textbf{Workflow Overview.} We propose agentic workflows for visually driven and speech-driven rough-cut, respectively. They transform the input materials into structured cut plans for subsequent
evaluation. For the visual setting, a natural end-to-end solution is to feed sampled frames from all source videos into a LVLM in a single call. However, the resulting long multi-video context requires extensive visual-token processing, increasing inference latency. Reasoning over all visual evidence is also difficult and may lead to unstable outputs. To address these issues, our visual workflow consists of a \textit{Video-Understanding Agent} and a \textit{Cut-Planning Agent}. The former processes each source video independently to detect candidate highlights and provide corresponding reasons, highlight-level visual
descriptions, and a video-level content summary. The latter uses these intermediate outputs to select the final highlight segments and determine the source-video order. For speech-driven rough-cut, we propose a \textit{Transcript-Cleanup Agent} that processes each complete ASR transcript and outputs structured cleanup decisions, reasons for each decision, and the cleaned transcript. 

\par \textbf{Video-Understanding Agent.} To reduce the difficulty of reasoning \cite{wei2022chain} over long multi-video contexts, the Video-Understanding Agent processes each source video independently. It analyzes frames sampled at 1 FPS, with frame timestamps included in the model input. The agent summarizes the overall video content and detects candidate highlights. For each candidate, it explains why the moment is worth considering and provides a description of its content. The output contains a video-level summary and candidate highlight records with start and end times, reasons, and descriptions. After the output is parsed, a representative key frame is extracted from the midpoint of each candidate highlight. These summaries, records, and key frames provide compact visual-semantic inputs for the subsequent Cut-Planning Agent.

\par \textbf{Cut-Planning Agent.} After all source videos are processed independently and in parallel by the \textit{Video-Understanding Agent}, their video-level summaries and candidate-highlight records are concatenated into a single input for the \textit{Cut-Planning Agent}. Each candidate highlight is labeled with its source-video identifier, such as \texttt{video\_0}, and includes its start and end timestamps, reason, description, and representative key frame. When provided, the user prompt is appended to the end of this input. The agent selects the final highlight segments from the candidates and determines the source-video order. It also adjusts segment boundaries when a target duration is specified. The final output is organized into a structured cut plan that records the source-video order and the selected highlight segments with their start and end timestamps.

\par \textbf{Transcript-Cleanup Agent.} Guided by the system prompt, the \textit{Transcript-Cleanup Agent} processes each complete ASR transcript in context to perform filler removal, expression deduplication, and low-value content trimming while preserving the original expression order and sentence structure. It outputs structured cleanup records together with the cleaned transcript. Each record specifies the content to remove, the content to retain when applicable, the cleanup type, and a corresponding reason. 

\vspace{-0.05 cm}
\vspace{0.05 cm}
\section{Experiments}

\par \textbf{Experimental Overview.}
Using the proposed agentic workflows, we evaluate representative LVLMs on visually driven and speech-driven rough-cut. Our experiments report annotation-based results for both settings, human-aligned evaluation of visual cuts, and ablations of key workflow designs.

\subsection{Experimental Setup}

\par \textbf{Evaluated Models.}
To evaluate current LVLMs on automatic rough-cut, we select eight representative models from multiple model families. The evaluated models are grouped as follows: (1) Gemini-family models \cite{team2023gemini}, including Gemini 2.5 Pro and Gemini 3.0 Pro; (2) GPT-family models \cite{achiam2023gpt}, including GPT-5.4 and GPT-5.5; (3) Qwen-family models \cite{yang2025qwen3}, including Qwen3.5-Plus and Qwen3.7-Plus; and (4) other recent multimodal models, including GLM-5V-Turbo \cite{hong2026glm} and Seed-2.0-Lite \cite{guo2025seed1}. Within each setting, all models are evaluated using the same workflow and evaluation protocol.

\par \textbf{Evaluation Sets and Execution Protocol.}
For the visual setting, we evaluate models on 101 cases: \textit{Visual Set-1} contains 62 general rough-cut cases, while \textit{Visual Set-2} contains 39 requirement-constrained cases with user instructions such as target duration, source-video order, or highlights to retain or remove. The resulting metrics are averaged. For human-aligned evaluation, predicted cut plans are rendered into videos and scored by the human-aligned agentic evaluator. For the speech-driven setting, we evaluate models on 84 speech-driven cases. Each complete transcript is processed by the \textit{Transcript-Cleanup Agent}; the resulting cleanup decisions and cleaned transcript are then compared with the span annotations. For speech-driven evaluation, we run each model ten times to reduce dependence on a single run. We sum the TP, FP, and FN counts across the ten runs and compute precision, recall, and F1 from these totals. RM$\rightarrow$R success is likewise calculated from the summed successful-pair and evaluated annotated-pair counts.

\par \textbf{Evaluation Metrics.}
We use complementary metrics to evaluate different aspects of predicted cut plans. For the visual setting, we evaluate predicted highlights using IoU-based \textit{precision}, \textit{recall}, and \textit{F1}, and report \textit{average IoU} for their temporal overlap with annotated highlights. Source-video ordering is evaluated with \textit{Kendall's $\tau$}. In addition to these annotation-based metrics, we use a Gemini 3.0 Pro-based human-aligned agentic evaluator to assess rendered cuts, and compare its scoring trend with professional scores. For the speech-driven setting, we evaluate cleanup decisions using the IM, RM$\rightarrow$R, and D labels defined in \textbf{Sec.~\ref{sec:data-construction}}. We report span-level \textit{precision}/\textit{recall}/\textit{F1} for IM, word-level \textit{precision}/\textit{recall}/\textit{F1} for D, and pair-level success rate for RM$\rightarrow$R.

\subsection{Visual Annotation-Based Results}\label{sec:visual-reference-results}

\par \textbf{Overall Performance.}
Table~\ref{tab:visual-reference-results} reports the annotation-based results for the visual setting. Overall, rough-cut remains challenging for current LVLMs, and no single model dominates all metrics. On Set-1, the best F1 and average IoU are 0.526 and 0.441, respectively; on Set-2, they are 0.343 and 0.358. These results indicate that Set-2 is more challenging for highlight detection and selection. In contrast, every model achieves a higher Kendall's
  $\tau$ on Set-2 than on Set-1.

\par \textbf{Highlight Detection and Selection.}
The precision-recall results reveal different highlight selection behaviors across models. For example, GPT-5.5 and Seed-2.0-Lite obtain high recall on Set-1, but their precision and average IoU are lower. This indicates recall-oriented behavior: their predictions cover a larger proportion of annotated highlights but include more unmatched segments and achieve lower overall temporal overlap. In contrast, Gemini-family models provide a more balanced selection behavior. Gemini 3.0 Pro achieves the best F1 on both Set-1 and Set-2, suggesting that it better balances coverage of annotated highlights and suppression of irrelevant segments. A similar mismatch appears in average IoU. Qwen3.5-Plus obtains the best average IoU on Set-1 but does not achieve the best F1. Together, these results show that highlight selection is not a single capability: coverage, selectivity, and temporal overlap can vary independently across models.

\par \textbf{Source-video Ordering.}
Beyond selecting highlight segments, rough-cut also requires models to determine a source-video order. The ordering results show a different capability profile from highlight selection and temporal localization. Seed-2.0-Lite achieves the strongest Kendall's $\tau$ on Set-2, and Qwen3.7-Plus also shows competitive ordering performance, but both models have lower highlight selection and localization scores than Gemini-family models. Conversely, Gemini-family models achieve stronger F1 and average IoU, but do not dominate source-video ordering across both sets. These results show that stronger highlight detection and selection does not necessarily translate into better source-video ordering.

\subsection{Human-Aligned Visual Evaluation}\label{sec:human-aligned-visual-evaluation}
\begin{table*}[t!]
\centering
\normalsize
\renewcommand{\arraystretch}{1.05}
\setlength{\tabcolsep}{2.0pt}
\begin{tabularx}{\textwidth}{l|*{4}{>{\centering\arraybackslash}X}|*{5}{>{\centering\arraybackslash}X}}
\hline
\multirow{2}{*}{\textbf{Model\textbackslash Metric}} & \multicolumn{4}{c|}{\textbf{\textit{Visual Set-1}}} & \multicolumn{5}{c}{\textbf{\textit{Visual Set-2}}} \\
\cline{2-5} \cline{6-10}
& {\footnotesize\textbf{Ordering}} & {\footnotesize\textbf{Redundancy}} & {\footnotesize\textbf{Overall}} & {\footnotesize\mbox{\textbf{Human Score}}} & {\footnotesize\textbf{Ordering}} & {\footnotesize\textbf{Requirement}} & {\footnotesize\textbf{Redundancy}} & {\footnotesize\textbf{Overall}} & {\footnotesize\mbox{\textbf{Human Score}}} \\
\hline
\rowcolor[rgb]{0.969,0.980,1.000}Gemini 2.5 Pro & \textbf{3.33} & \underline{2.52} & \underline{3.11} & \underline{3.46} & \underline{3.66} & \textbf{4.31} & \textbf{2.97} & \textbf{3.60} & \textbf{3.76} \\
\rowcolor[rgb]{0.969,0.980,1.000}Gemini 3.0 Pro & \underline{3.30} & \textbf{2.58} & \textbf{3.13} & \textbf{3.52} & 3.46 & 4.09 & \underline{2.87} & 3.42 & 3.50 \\
\hline
\rowcolor[rgb]{0.973,1.000,0.969}GPT-5.4 & 3.15 & 2.07 & 2.82 & 2.70 & \textbf{3.69} & 4.15 & 2.68 & 3.47 & 3.60 \\
\rowcolor[rgb]{0.973,1.000,0.969}GPT-5.5 & \underline{3.30} & 2.05 & 2.88 & 2.88 & 3.43 & 4.16 & 2.68 & 3.45 & 3.42 \\
\hline
\rowcolor[rgb]{1.000,0.992,0.957}Qwen3.5-Plus & 3.18 & 2.32 & 2.95 & 3.22 & 3.35 & 4.14 & 2.68 & 3.36 & 3.35 \\
\rowcolor[rgb]{1.000,0.992,0.957}Qwen3.7-Plus & 2.99 & 2.24 & 2.86 & 2.82 & 3.53 & \underline{4.21} & 2.80 & \underline{3.49} & \underline{3.68} \\
\hline
\rowcolor[rgb]{0.984,0.973,1.000}GLM-5V-Turbo & 3.19 & 2.24 & 2.90 & 3.06 & 3.26 & 4.03 & 2.63 & 3.32 & 3.26 \\
\rowcolor[rgb]{0.984,0.973,1.000}Seed-2.0-Lite & 3.04 & 2.07 & 2.73 & 2.56 & 3.38 & 3.96 & 2.62 & 3.27 & 3.08 \\
\hline
\end{tabularx}
\vspace{0em}
\caption{Human-aligned agentic evaluation for rendered cuts (Sec.~\ref{sec:human-aligned-visual-evaluation}). The agentic evaluator scores ordering and redundancy control, together with requirement following for \textit{Visual Set-2}, on a 1--5 scale. \textit{Overall} denotes the visual quality score assigned by jointly considering the evaluated dimensions. The best and second-best scores are shown in bold and underlined, respectively.}
\label{tab:visual-agentic-results}
\vspace{-0.7 cm}
\end{table*}

\begin{table}[t!]
\centering
\footnotesize
\renewcommand{\arraystretch}{1.05}
\setlength{\tabcolsep}{1.5pt}
\begin{tabularx}{\columnwidth}{l|*{3}{>{\centering\arraybackslash}X}|>{\centering\arraybackslash}X|*{3}{>{\centering\arraybackslash}X}}
\hline
\multirow{2}{*}{\textbf{Model\textbackslash Metric}} & \multicolumn{3}{c|}{\textbf{\textit{IM}}} & \textbf{\textit{RM$\rightarrow$R}} & \multicolumn{3}{c}{\textbf{\textit{D}}} \\
\cline{2-4} \cline{5-5} \cline{6-8}
& \textbf{P} & \textbf{R} & \textbf{F1} & \textbf{Succ.} & \textbf{P} & \textbf{R} & \textbf{F1} \\
\hline
\rowcolor[rgb]{0.969,0.980,1.000}Gemini 2.5 Pro & 0.430 & 0.891 & 0.581 & 0.491 & 0.707 & 0.609 & \underline{0.654} \\
\rowcolor[rgb]{0.969,0.980,1.000}Gemini 3.0 Pro & 0.416 & \underline{0.898} & 0.569 & 0.497 & 0.692 & 0.608 & 0.648 \\
\hline
\rowcolor[rgb]{0.973,1.000,0.969}GPT-5.4 & 0.332 & 0.837 & 0.475 & 0.357 & 0.651 & \textbf{0.693} & \textbf{0.671} \\
\rowcolor[rgb]{0.973,1.000,0.969}GPT-5.5 & 0.403 & \textbf{0.913} & 0.559 & 0.340 & 0.670 & \underline{0.610} & 0.639 \\
\hline
\rowcolor[rgb]{1.000,0.992,0.957}Qwen3.5-Plus & 0.507 & 0.786 & 0.616 & \underline{0.535} & 0.733 & 0.444 & 0.553 \\
\rowcolor[rgb]{1.000,0.992,0.957}Qwen3.7-Plus & \underline{0.612} & 0.661 & \underline{0.635} & 0.485 & \textbf{0.827} & 0.448 & 0.581 \\
\hline
\rowcolor[rgb]{0.984,0.973,1.000}GLM-5V-Turbo & 0.533 & 0.769 & 0.630 & \textbf{0.537} & 0.745 & 0.428 & 0.543 \\
\rowcolor[rgb]{0.984,0.973,1.000}Seed-2.0-Lite & \textbf{0.691} & 0.650 & \textbf{0.670} & 0.446 & \underline{0.750} & 0.104 & 0.183 \\
\hline
\end{tabularx}
\vspace{0 cm}
\caption{Annotation-based results for the speech-driven setting (Sec.~\ref{sec:speech-driven-reference-results}). The best and second-best scores
are shown in bold and underlined, respectively.}
\label{tab:speech-reference-results}
\vspace{-1.1 cm}
\end{table}

For human-aligned visual evaluation, we render each predicted cut plan into a video and score it with the evaluator described in Sec.~\ref{sec:evaluation-protocols}. The evaluator assigns each rendered cut an integer score from 1 to 5 for each applicable evaluation dimension, together with an overall visual quality score based on these dimensions. Overall visual quality is assessed holistically rather than as an arithmetic mean of dimension scores; requirement following is excluded when no user requirements are provided. It evaluates each rendered cut three times, and the resulting scores are averaged. Model-level scores are then averaged over all rendered cuts in each evaluation set. In the user study, three professional annotators independently assign every rendered cut an integer overall visual quality score from 1 to 5 and provide a corresponding reason. For each model and evaluation set, the Human Score is obtained by averaging these ratings across annotators and rendered cuts.

\begin{figure}[t!]
    \centering
    \includegraphics[width=\linewidth]{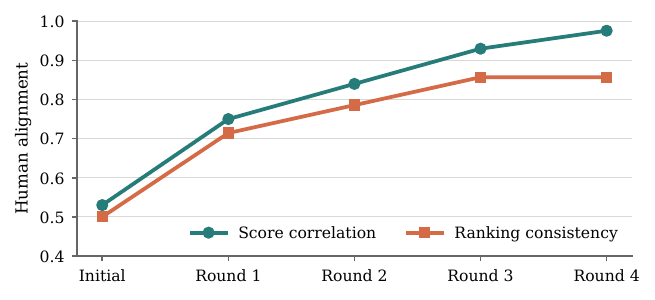}
    \vspace{-0.8 cm}
    \caption{Agreement between the agentic evaluator and professional assessments over four feedback rounds (Sec.~\ref{sec:human-aligned-visual-evaluation}).}
    \label{fig:evaluator-feedback}
    \vspace{-0.5 cm}
\end{figure}

\par \textbf{Feedback Alignment.}
The professional scores and reasons remain fixed throughout the four feedback rounds. In each round, we compare the evaluator's scores and reasons with those provided by professionals and refine its system prompt accordingly. At each round, the evaluator score is first averaged over three evaluations of each rendered cut and then over all rendered cuts from the LVLM. We compute \textit{score correlation} between the evaluator scores and the professional scores using Pearson correlation, which measures whether LVLMs receiving higher professional scores also receive higher evaluator scores. Separately, we rank the eight LVLMs once by their evaluator scores and once by their professional scores. We then compute \textit{ranking consistency} between these two rankings using Kendall's $\tau$. As shown in Fig.~\ref{fig:evaluator-feedback}, score correlation improves overall from 0.53 to 0.98, while ranking consistency improves from 0.50 to 0.86. Most improvement occurs in the early feedback rounds, followed by smaller changes as the evaluator approaches stable agreement with professional assessments.

\par \textbf{Result Analysis.}
As shown in Table~\ref{tab:visual-agentic-results}, Gemini-family models achieve the strongest overall results. Gemini 3.0 Pro obtains the highest overall score on Set-1 (3.13), while Gemini 2.5 Pro performs best on Set-2 (3.60). On Set-2, requirement-following scores remain relatively high across all models, whereas redundancy scores range from 2.62 to 2.97. This suggests that current models follow explicit user requirements more reliably than they control redundant content in the rendered cut. Models with higher evaluator scores generally also receive higher professional scores, although the rankings produced by the two evaluations differ slightly.
\subsection{Speech-Driven Annotation-Based Results}\label{sec:speech-driven-reference-results}

\par \textbf{Main Results.}
Table~\ref{tab:speech-reference-results} reports the annotation-based results for the speech-driven setting. No model performs best across all speech-cleanup categories. Seed-2.0-Lite achieves the highest IM F1, followed by Qwen3.7-Plus. Their leading F1 scores indicate stronger performance in identifying and removing short low-content markers. In contrast, RM$\rightarrow$R remains challenging across models, with the best pair-level success rate reaching only 0.537. For low-value content trimming, GPT-5.4 obtains the highest D F1 and recall, while Gemini-family models also perform competitively. Seed-2.0-Lite shows a different pattern, achieving high D precision but low recall. This pattern indicates that Seed-2.0-Lite tends to retain more content: it makes fewer incorrect deletions but fails to remove many spans marked by annotators.
  
\begin{table*}[t!]
\centering
\normalsize
\renewcommand{\arraystretch}{1.05}
\setlength{\tabcolsep}{1.2pt}
\begin{tabularx}{\textwidth}{l|*{5}{>{\centering\arraybackslash}X}|*{5}{>{\centering\arraybackslash}X}}
\hline
\multirow{2}{*}{\textbf{Input}} & \multicolumn{5}{c|}{\textbf{\textit{Visual Set-1}}} & \multicolumn{5}{c}{\textbf{\textit{Visual Set-2}}} \\
\cline{2-6} \cline{7-11}
& {\footnotesize\textbf{Precision}} & {\footnotesize\textbf{Recall}} & {\footnotesize\textbf{F1}} & {\footnotesize\textbf{Avg. IoU}} & {\footnotesize\textbf{Kendall's $\tau$}} & {\footnotesize\textbf{Precision}} & {\footnotesize\textbf{Recall}} & {\footnotesize\textbf{F1}} & {\footnotesize\textbf{Avg. IoU}} & {\footnotesize\textbf{Kendall's $\tau$}} \\
\hline
\textbf{\textit{Time + Frame + Desc}} & \textbf{0.490} & \textbf{0.568} & \textbf{0.526} & \underline{0.440} & \textbf{0.565} & \underline{0.345} & \textbf{0.342} & \underline{0.343} & \underline{0.358} & \textbf{0.692} \\
Time + Frame & 0.474 & \underline{0.566} & \underline{0.516} & \underline{0.440} & 0.518 & \textbf{0.348} & \underline{0.340} & \textbf{0.344} & \textbf{0.365} & 0.474 \\
Time + Desc & \underline{0.480} & 0.556 & 0.515 & \textbf{0.450} & \underline{0.556} & 0.295 & 0.303 & 0.299 & \underline{0.358} & \underline{0.618} \\
\hline
\end{tabularx}
\vspace{0 cm}
\caption{Ablation results of input components for the \textit{Cut-Planning Agent} (Sec.~\ref{sec:ablation-studies}). Time, Frame, and Desc denote timestamps, representative key frames, and segment descriptions, respectively. We use \textit{Time + Frame + Desc} as the default setting. }
\label{tab:visual-arrange-input-ablation}
% \vspace{-0.8 cm}
\vspace{-0.5 cm}
\end{table*}

\par \textbf{Capability Dissociation.}
The speech results further show that transcript cleanup is not a single capability. A model that performs well on filler removal may not perform equally well on low-value content trimming. For example, Seed-2.0-Lite obtains the best IM F1 but the weakest D F1, whereas GPT-5.4 shows the opposite pattern. This reveals a dissociation between local filler removal and low-value content judgment: performance in one cleanup category does not reliably predict performance in another.

\subsection{Ablation Studies}\label{sec:ablation-studies}

\par \textbf{Ablation Scope.}
We study two visual workflow choices: the frame sampling rate used by the \textit{Video-Understanding Agent} and the input representation used by the \textit{Cut-Planning Agent}. The former controls how densely each source video is sampled into timestamped frames, affecting both the available visual evidence and inference cost. We further examine the reasoning strategy of the \textit{Transcript-Cleanup Agent}. Additional ablations on the overall workflow design are provided in Appendix. All ablation experiments are conducted using Gemini 3.0 Pro.

\begin{table*}[t!]
\centering
\normalsize
\renewcommand{\arraystretch}{1.05}
\setlength{\tabcolsep}{1.2pt}
\begin{tabularx}{\textwidth}{l|*{5}{>{\centering\arraybackslash}X}|*{5}{>{\centering\arraybackslash}X}}
\hline
\multirow{2}{*}{\textbf{FPS}} & \multicolumn{5}{c|}{\textbf{\textit{Visual Set-1}}} & \multicolumn{5}{c}{\textbf{\textit{Visual Set-2}}} \\
\cline{2-6} \cline{7-11}
& {\footnotesize\textbf{Precision}} & {\footnotesize\textbf{Recall}} & {\footnotesize\textbf{F1}} & {\footnotesize\textbf{Avg. IoU}} & {\footnotesize\textbf{Kendall's $\tau$}} & {\footnotesize\textbf{Precision}} & {\footnotesize\textbf{Recall}} & {\footnotesize\textbf{F1}} & {\footnotesize\textbf{Avg. IoU}} & {\footnotesize\textbf{Kendall's $\tau$}} \\
\hline
0.5 & 0.466 & 0.537 & 0.499 & 0.436 & 0.547 & 0.284 & 0.289 & 0.286 & 0.348 & 0.651 \\
\textbf{\textit{1.0}} & \textbf{0.490} & \underline{0.568} & \underline{0.526} & \underline{0.440} & \underline{0.565} & \underline{0.345} & \underline{0.342} & \underline{0.343} & \textbf{0.358} & \underline{0.692} \\
2.0 & \underline{0.486} & \textbf{0.576} & \textbf{0.527} & \textbf{0.445} & \textbf{0.614} & \textbf{0.357} & \textbf{0.364} & \textbf{0.360} & \underline{0.357} & \textbf{0.732} \\
\hline
\end{tabularx}
% \vspace{0.2em}
\caption{Ablation results of FPS in the \textit{Video-Understanding Agent} (Sec.~\ref{sec:ablation-studies}). We use 1 FPS as the default frame sampling rate.}
\label{tab:visual-fps-ablation}
% \vspace{-0.8 cm}
\vspace{-0.6 cm}
\end{table*}

\par \textbf{Frame Sampling Rate.}
We study the effect of frame sampling rate in the \textit{Video-Understanding Agent} using Gemini 3.0 Pro. As shown in Table~\ref{tab:visual-fps-ablation}, 0.5 FPS leads to consistently lower performance than 1 FPS and 2 FPS, especially on Set-2, indicating that sparse sampling misses useful visual evidence for highlight detection and selection. Increasing the sampling rate to 2 FPS brings higher recall and stronger source-video ordering, but it does not uniformly improve all metrics: 1 FPS achieves the best precision on Set-1 and the best average IoU on Set-2. Since denser sampling increases the number of visual inputs, we use 1 FPS as the default setting. It achieves competitive performance across both sets while requiring fewer input frames than 2 FPS.

\par \textbf{Cut-Planning Input Representation.}
Table~\ref{tab:visual-arrange-input-ablation} shows that combining timestamps, key frames, and segment descriptions gives the most balanced performance across the two visual sets. Removing descriptions retains comparable highlight selection scores but reduces Kendall's $\tau$ on Set-2 from 0.692 to 0.474, indicating that descriptions provide important semantic evidence for source-video ordering. Removing key frames reduces Set-2 precision, recall, F1, and ordering, suggesting that direct visual evidence also supports highlight selection and planning. Together, descriptions and key frames provide complementary evidence for the Cut-Planning Agent.

\begin{table}[t!]
\centering
\footnotesize
\renewcommand{\arraystretch}{1.05}
\setlength{\tabcolsep}{1.5pt}
\begin{tabularx}{\columnwidth}{l|*{3}{>{\centering\arraybackslash}X}|>{\centering\arraybackslash}X|*{3}{>{\centering\arraybackslash}X}}
\hline
\multirow{2}{*}{\textbf{Strategy\textbackslash Metric}} & \multicolumn{3}{c|}{\textbf{\textit{IM}}} & \textbf{\textit{RM$\rightarrow$R}} & \multicolumn{3}{c}{\textbf{\textit{D}}} \\
\cline{2-4} \cline{5-5} \cline{6-8}
& \textbf{P} & \textbf{R} & \textbf{F1} & \textbf{Succ.} & \textbf{P} & \textbf{R} & \textbf{F1} \\
\hline
\textbf{\textit{CoT zero-shot}} & 0.416 & \underline{0.898} & 0.569 & \textbf{0.497} & 0.692 & \underline{0.608} & \underline{0.648} \\
CoT one-shot & 0.385 & \textbf{0.910} & 0.542 & \underline{0.491} & 0.671 & \textbf{0.688} & \textbf{0.680} \\
w/o CoT zero-shot & \underline{0.466} & 0.831 & \underline{0.597} & 0.451 & \underline{0.693} & 0.388 & 0.497 \\
w/o CoT one-shot & \textbf{0.512} & 0.741 & \textbf{0.605} & 0.398 & \textbf{0.702} & 0.332 & 0.451 \\
\hline
\end{tabularx}
\vspace{0 cm}
\caption{Ablation results of reasoning strategies for the \textit{Transcript-Cleanup Agent} (Sec.~\ref{sec:ablation-studies}). We use \textit{CoT zero-shot} as the default setting.}
\label{tab:speech-prompt-ablation}
\vspace{-1 cm}
\end{table}

\par \textbf{Transcript-Cleanup Reasoning Strategy.}
Table~\ref{tab:speech-prompt-ablation} shows that reasoning strategy affects speech-cleanup differently. CoT improves low-value content trimming under both zero-shot and one-shot prompting, with the one-shot variant achieving the highest D F1. It also improves RM$\rightarrow$R handling, whereas removing CoT produces higher IM F1. These results suggest that explicit reasoning is more useful for low-value content trimming and expression deduplication, while direct prediction is more effective for filler removal.

\section{Discussion}

\par \textbf{Implications for Rough-Cut Systems.}
Our results show that rough-cut is not a monolithic capability: LVLMs perform unevenly across highlight detection and selection, source-video ordering, adherence to user requirements, and spoken-content refinement. For visual rough-cut, future methods need stronger long-context reasoning for cut planning to identify highlight moments, distinguish event progression from repeated actions, and use temporal and spatial relations for source-video ordering. For ASR-transcript-based rough-cut, models need multi-granularity reasoning to distinguish fillers, speech repairs, and longer low-value passages from content-carrying speech. Training data and supervision should therefore emphasize structured cleanup decisions and their relations rather than isolated content recognition.

\par \textbf{Evaluation Boundaries.}
Our agentic workflows support cross-model comparability and reliable execution rather than maximizing each model's attainable performance. Alternative workflows that process all source videos jointly or combine understanding and planning in a single request can improve some metrics, but not all model APIs can reliably handle the larger visual contexts required by these workflows (Appendix). The annotations provide a consistent reference, but other cut plans may also be acceptable. The human-aligned agentic evaluator broadens evaluation beyond annotation agreement by assessing rendered-cut quality. Its model-level trends are broadly consistent with the professional user study, supporting its use as a scalable complement to human assessment.

\par \textbf{Ethics, Release, and Future Scope.}
RoughCut-Bench is curated from real-world media creation data, and all released data are screened to remove sensitive personal information and platform-specific identifiers. We will release the cleared source videos and ASR transcripts together with professional annotations, metadata, workflow and evaluator prompts, evaluation scripts, and rendering tools, subject to the corresponding usage permissions. The current benchmark studies visual planning and transcript cleanup separately, while both produce decisions that can be translated into cuts on the video timeline. However, the current speech setting does not use audio-visual cues such as prosody, emotion, or speaker interaction. Future versions will extend to joint audio-visual rough-cut and broader user requirements. Beyond evaluation, the accompanying workflows and evaluation criteria can support quality filtering of separately collected rough-cut examples, facilitating the construction of post-training data. We also plan to develop a rough-cut-specific omni-modal model for generating executable cut plans from multimodal rough-cut inputs.

\section{Conclusion}

In this work, we present \textbf{\textit{RoughCut-Bench}}, a benchmark for automatic video rough-cut. RoughCut-Bench covers visually driven rough-cut over video cases and speech-driven rough-cut over ASR transcripts, providing professionally annotated cases, proposed agentic workflows, and complementary evaluation protocols. For the visual setting, the benchmark evaluates highlight detection and selection, source-video ordering, adherence to user requirements when provided, and rendered-cut quality; for the speech-driven setting, it evaluates filler removal, expression deduplication, and low-value content trimming over ASR transcripts. Experiments with representative LVLMs using the proposed agentic workflows show that current LVLMs still struggle to produce expert-aligned cut plans across both settings. We hope RoughCut-Bench will help establish automatic video rough-cut as a testbed for media creation.

\textbf{Contributions and Support.} We sincerely thank Huiqi Zhang for her valuable contributions to this benchmark, including dataset procurement, coordination of annotation efforts, review of annotation results, and overall coordination of human evaluation.

\textbf{Acknowledgements.} This work was supported by the Fundamental and Interdisciplinary Disciplines Breakthrough Plan of the Ministry of Education of China (No. JYB2025XDXM103).

\bibliographystyle{ACM-Reference-Format}
\bibliography{main}
\appendix
\twocolumn[{%
  \begin{minipage}{\textwidth}
\centering
\small
\renewcommand{\arraystretch}{1.05}
\setlength{\tabcolsep}{0pt}
\begin{tabularx}{\textwidth}{l|*{5}{>{\centering\arraybackslash}X}|*{5}{>{\centering\arraybackslash}X}}
\hline
\multirow{2}{*}{\textbf{Workflow\textbackslash Metric}} & \multicolumn{5}{c|}{\textbf{\textit{Visual Set-1}}} & \multicolumn{5}{c}{\textbf{\textit{Visual Set-2}}} \\
\cline{2-6} \cline{7-11}
& {\footnotesize\textbf{Precision}} & {\footnotesize\textbf{Recall}} & {\footnotesize\textbf{F1}} & {\footnotesize\textbf{Avg. IoU}} & {\footnotesize\textbf{Kendall's $\tau$}} & {\footnotesize\textbf{Precision}} & {\footnotesize\textbf{Recall}} & {\footnotesize\textbf{F1}} & {\footnotesize\textbf{Avg. IoU}} & {\footnotesize\textbf{Kendall's $\tau$}} \\
\hline
\textbf{Decoupled} & 0.490 & \textbf{0.568} & \textbf{0.526} & 0.440 & 0.565 & 0.345 & 0.342 & 0.343 & 0.358 & 0.692 \\
Decoupled-joint understanding & \textbf{0.501} & \underline{0.532} & \underline{0.516} & \textbf{0.465} & \underline{0.569} & \textbf{0.387} & \textbf{0.360} & \textbf{0.373} & \underline{0.391} & \textbf{0.701} \\
Unified w/ understanding & \underline{0.495} & 0.523 & 0.509 & \underline{0.453} & \textbf{0.612} & \underline{0.375} & \underline{0.352} & \underline{0.363} & \textbf{0.400} & \underline{0.695} \\
\hline
\end{tabularx}
\vspace{0.2em}
\captionof{table}{Workflow design ablation with Gemini 3.0 Pro. The boldfaced row label denotes the default setting used in the main experiments.}
\label{tab:supp-workflow-ablation}
\vspace{-0.8em}
\end{minipage}

}]
% \clearpage

\section{Additional Details}

\subsection{Implementation and Execution Details}
\label{sec:supp-implementation-details}

Model outputs are normalized and validated before evaluation, with missing or malformed results regenerated through the repair procedure. For model endpoints subject to a 120-image limit, representative frames are retained for the 120 longest candidate segments, while the timestamps and descriptions of all candidates remain available to the \textit{Cut-Planning Agent}. Additional endpoint configurations and validation rules will be provided in the released implementation.

\subsection{Workflow Design Ablation}

We further study how different workflow designs affect visually driven rough-cut performance. This ablation is conducted with Gemini 3.0 Pro to control for model differences. The goal is to examine how multi-video understanding and cut planning should be organized: whether the model should first understand each source video separately, jointly understand all source videos in a case, or directly generate the final cut plan in a unified manner.

We compare three workflow variants. \textbf{Decoupled} is our default workflow, where each source video is first processed independently by the Video-Understanding Agent, and the resulting candidate highlights are then passed to the Cut-Planning Agent. \textbf{Decoupled-joint understanding} keeps the two-stage structure, but performs the understanding stage over all source videos in the same case jointly, so that candidate highlights are extracted with access to cross-video context. \textbf{Unified w/ understanding} feeds all source videos in a case to a single model call and explicitly asks the model to output both intermediate understanding results and the final cut plan.

Table~\ref{tab:supp-workflow-ablation} reports the results. Compared with the default decoupled workflow, joint understanding improves temporal localization, increasing average IoU from 0.440 to 0.465 on Set-1 and from 0.358 to 0.391 on Set-2. This suggests that the understanding stage benefits from cross-video context: even though highlights are selected within each source video, knowing the other source videos in the same case helps the model judge which moments are more informative and how their boundaries should be localized.

The unified workflow improves average IoU and source-video ordering over the default workflow on both sets, but its recall and F1 are lower on Set-1. This result indicates that combining understanding and planning in a single request can strengthen some aspects of global planning without uniformly improving highlight detection and selection.

Overall, this ablation reveals a trade-off between joint reasoning capacity and workflow robustness. Joint or unified reasoning can improve performance for strong long-context LVLMs, but these settings require processing many visual inputs in a single call and are more sensitive to context length, image-count limits, and structured-output errors. In practice, not all model families or API endpoints can accept such large image contexts, and some fail directly under these inputs. We therefore use the decoupled workflow as the default protocol for broad model comparison, prioritizing cross-model comparability and robustness rather than maximizing the absolute performance of each individual model. The joint and unified variants are used as diagnostic ablations to understand the potential benefits and limitations of more integrated rough-cut workflows.

\end{document}